# KOAL: Knowledge-Driven Prostate Cancer Grading with Ordinal-Aware Learning

Zheng Guo[1†], Jiaqi Cui[1†], Haocheng Xiong[1], Jize Han[2], Bo Liu[1], Qianwen Zhang[3(✉)], Rui Chen[4(✉)], and Yan Wang[1(✉)]

[1] School of Computer Science, Sichuan University, China
wangyanscu@hotmail.com
[2] China Mobile (Suzhou) Software Technology Company Limited, China
[3] Shanghai Changhai Hospital, Naval Medical University, China
[4] Renji Hospital, Shanghai Jiao Tong University School of Medicine, China
drchenrui@foxmail.com

**Abstract.** Non-invasive prediction of Gleason Grade Group (GGG) in prostate cancer using multiparametric MRI (mpMRI) is clinically vital for reducing unnecessary biopsies. Existing GGG prediction methods face two major limitations. First, they often overlook non-image information critical for GGG prediction, including age, prostate-specific antigen (PSA), and expert priors embedded in radiology reports. Second, they tend to oversimplify GGG as flat categorical labels, failing to account for its intrinsic hierarchy of primary and secondary Gleason patterns. To this end, we propose a novel **K**nowledge-Driven **O**rdinal-**A**ware **L**earning (KOAL) framework with three synergistic modules. Specifically, the Clinical-Context Modulation (CCM) module uses clinical variables (e.g., age and PSA) to dynamically modulate discriminative image representations. The Knowledge-Guided Prototype Alignment (KGPA) module leverages an LLM to extract group-specific expert knowledge from training radiology reports and clinical guidelines, producing offline semantic anchors describing grade-specific radiological findings without requiring patient-specific reports at inference. Through prototype contrastive alignment, patient-specific mpMRI representations are matched with these anchors to promote pathology-aligned representation learning. The Hierarchical Ordinal-aware Constraints (HOC) module decouples primary and secondary Gleason pattern prediction and maps their probabilistic outputs to GGG via a Differentiable Bio-logic Mapping Layer (DBML), ensuring pathological grading consistency. Experiments on public PI-CAI and in-house datasets demonstrate that KOAL outperforms state-of-the-art methods. Code is available at: https://github.com/Gother-GZ/KOAL.

## 1 Introduction

Prostate cancer (PCa) is one of the most prevalent malignancies in men worldwide [1]. In clinical practice, tumor aggressiveness is central to diagnosis and treatment, and is usually evaluated using the Gleason grading system [2]. This system assigns grades to the dominant (i.e., primary) and the next-most frequent (i.e., secondary) patterns, which together form the Gleason score (e.g., 3+3 to 5+5) [2]. Given the established association between Gleason score and patient prognosis, the International Society of Urological Pathology (ISUP) further introduced a simplified grading system, i.e., the Gleason Grade Group (GGG, ranging from 1 to 5) [3]. However, as a pathological gold standard, obtaining GGG typically necessitates invasive biopsy or surgical resection, which is associated with procedure-related complications [4]. Moreover, inconsistencies often exist between biopsy-based GGG and whole-mount pathological analysis [5]. Therefore, reliable non-invasive GGG prediction is highly desirable for reducing unnecessary biopsies and supporting clinical decision-making.

† These authors contributed equally.

In clinic, multiparametric Magnetic Resonance Imaging (mpMRI), including T2-weighted imaging (T2), diffusion-weighted imaging (DWI), and apparent diffusion coefficient (ADC), has been widely adopted for non-invasive prostate cancer assessment [6,7]. Nevertheless, accurate prediction of histopathological GGG from mpMRI remains challenging due to intra-tumor heterogeneity, inter-observer variability, and subtle imaging-pathology correlations [8]. Early studies primarily employed radiomics-based approaches [9,10,11], which focused on extracting handcrafted features from mpMRI to establish a mapping between imaging phenotypes and GGG grading. However, these methods often require precise lesion segmentation and expert-defined feature engineering, which are labor-intensive and exhibit limited generalization capabilities across different centers. More recently, deep learning (DL) techniques have been explored to improve GGG prediction [12,13,14,15]. In addition, given the inherent ordinal nature of GGG (1-5), several studies have explored modeling ordinal relationships among grade groups. In this vein, Xu *et al.* [16] proposed a Poisson ordinal network for Gleason group estimation on MRI to improve consistency and robustness.

Despite their improvements, existing methods still face two key limitations. First, these approaches primarily rely on image-only mpMRI inputs for GGG prediction. This paradigm, however, often overlooks valuable expert knowledge conveyed in radiology reports, which reflect radiologists' case-by-case assessments that are difficult to infer from images alone. In addition, clinical variables such as age and prostate-specific antigen (PSA), which are well-established risk indicators for prostate cancer [20], are not involved, leading to compromised prediction accuracy. Second, existing methods either model GGG as flat categorical labels or, at best, as a single ordinal variable. Although these formulations recognize the ordinal nature of GGG, they treat it as an indivisible entity and fail to account for its intrinsic compositional structure, defined by primary and secondary Gleason patterns. As both patterns are ordinal, ignoring this fine-grained pattern-level structure results in an overly coarse modeling of disease severity.

Motivated to tackle these limitations, we propose a **K**nowledge-Driven **O**rdinal-**A**ware **L**earning (KOAL) framework for GGG prediction in prostate cancer. KOAL consists of three synergistic components, i.e., a Clinical-Context Modulation (CCM) module, a Knowledge-Guided Prototype Alignment (KGPA) module, and a Hierarchical Ordinal-aware Constraints (HOC) module. Specifically, the CCM module uses clinical variables (e.g., PSA and age) to condition a cross-attention mechanism, dynamically modulating image representations for patient-specific discrimination. Moreover, the KGPA module leverages a large language model (LLM) to extract grade group–specific knowledge from a large corpus of radiology reports. This process constructs a group-level knowledge bank that captures radiologists' population-level priors. From these priors, semantic anchors are derived to describe the radiological findings of each grade group. Through prototype contrastive alignment, patient-specific mpMRI representations are aligned with these semantic anchors, promoting pathology-aligned representation learning. In addition, to model the fine-grained ordinal structure of GGG, HOC decouples the GGG score prediction into primary and secondary Gleason patterns prediction, and then maps their probabilistic outputs to the final GGG via a Differentiable Bio-logic Mapping Layer (DBML) based on ISUP-defined combinatorial rules, thereby enforcing predictions consistent with clinical grading logic. Through the above designs, KOAL effectively exploits the potential of heterogeneous imaging (i.e., mpMRI) and non-imaging information (i.e., radiology reports and clinical variables) with fine-grained ordinal supervision for improved GGG prediction.

The main contributions of this paper are summarized as follows. (1) We propose KOAL, the first knowledge-driven ordinal-aware learning framework that integrates heterogeneous mpMRI, radiology reports, and clinical variables for GGG prediction in prostate cancer. (2) The KGPA module leverages an LLM to extract grade-group–specific knowledge from radiology reports and derives semantic anchors to align patient-specific mpMRI with population-level radiological priors. (3) The HOC module decouples primary and secondary Gleason patterns, mapping them to the final GGG using the DBML, thereby enforcing compositional ordinal constraints consistent with clinical grading. (4) Extensive experiments on the public PI-CAI dataset and an in-house dataset demonstrate that KOAL outperforms existing state-of-the-art approaches.

## 2 Methodology

The KOAL framework is illustrated in Fig. 1. It leverages three core modules, i.e., the Clinical-Context Modulation (CCM) module, the Knowledge-Guided Prototype Alignment (KGPA) module, and the Hierarchical Ordinal-aware Constraints (HOC) module, to integrate mpMRI, clinical variables, and radiology reports to improve non-invasive GGG prediction. The details of each module are described in the following subsections.

### 2.1 Clinical Context Modulation (CCM)

In clinical practice, doctors consider both mpMRI and clinical variables when assessing tumor aggressiveness and estimating Gleason grade. However, existing methods struggle to integrate mpMRI modalities with distinct focuses (i.e., anatomical structures from T2 and functional characteristics from DWI/ADC) and to fuse imaging with clinical variables. This results in suboptimal feature representations and limited prediction accuracy. As shown in Fig. 1(a), the CCM module tackles this challenge by combining channel-wise early fusion with clinically guided feature recalibration.

**MpMRI Image Encoding.** To exploit the complementarity of mpMRI, we adopt a channel-wise early fusion strategy. Let $\boldsymbol{S} = \{\boldsymbol{I}_m \in \mathbb{R}^{D\times H\times W}\}_{m\in M}$ denote a set of spa-tially aligned mpMRI volumes, where $M = \{T2, ADC, DWI\}$. First, we stack these volumes along the channel dimension to form $\boldsymbol{X} \in \mathbb{R}^{3\times D\times H\times W}$. Then, a 3D ResNet-50 backbone $\mathcal{B}(\cdot)$ [21] extracts visual feature maps $\boldsymbol{F}_v = \mathcal{B}(\boldsymbol{X})$. We further apply Global Average Pooling (GAP) and a projection head to obtain a pooled visual embedding $\boldsymbol{f}_v = \mathcal{P}_{map}\big(\mathrm{GAP}(\boldsymbol{F}_v)\big) \in \mathbb{R}^d$, where $\mathcal{P}_{map}(\cdot)$ denotes an additional mapping layer for reduction and normalization operations.

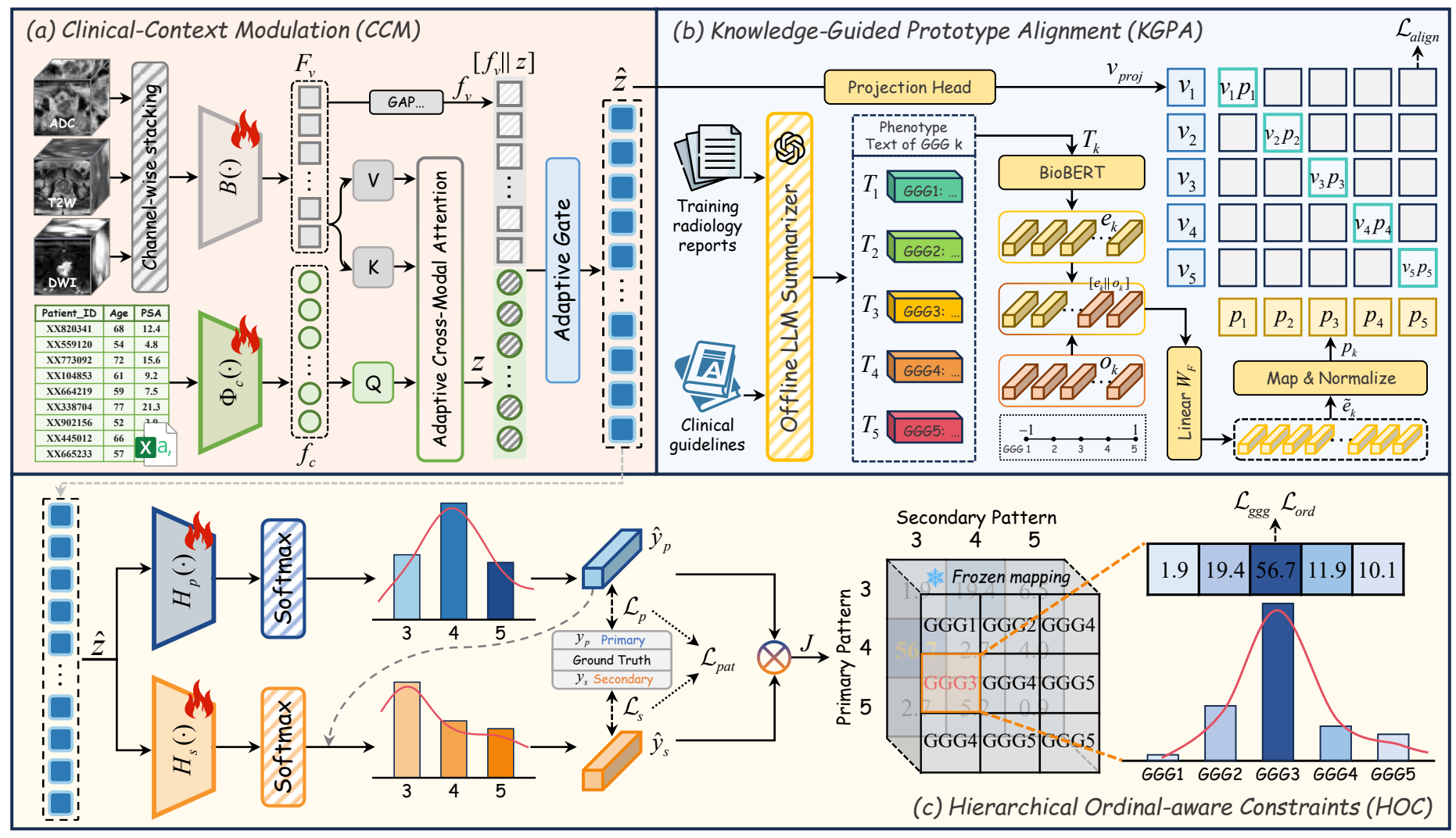

**Fig. 1.** Overview of the proposed KOAL framework. The architecture consists of three synergistic modules, including (a) Clinical-Context Modulation (CCM); (b) Knowledge-Guided Prototype Alignment (KGPA); and (c) Hierarchical Ordinal-aware Constraints (HOC).

**Clinical Variables Encoding.** Given a set of clinical variables $\boldsymbol{c} \in \mathbb{R}^L$, where $L$ is the length of the set, we employ a non-linear semantic mapping network $\Phi_c(\cdot)$ to transform $\boldsymbol{c}$ into a context-rich semantic embedding $\boldsymbol{f}_c$. This transformation is achieved through layer normalization LN$(\cdot)$ and multi-layer perceptron projection $\mathcal{P}_{mlp}(\cdot)$, formulated as $\boldsymbol{f}_c = \Phi_c(\boldsymbol{c}) = \mathcal{P}_{mlp}\Big(\delta\big(LN(\boldsymbol{c})\big)\Big) \in \mathbb{R}^d$, where $\delta$ denotes the GELU activation function. This mapping projects $\boldsymbol{f}_c$ into a shared semantic space with the image feature $\boldsymbol{f}_v$ for subsequent cross-modal interaction.

**Adaptive Cross-Modal Attention (ACMA).** To refine image representations using clinical variables, we develop the ACMA mechanism. Concretely, we treat $\boldsymbol{f}_c$ as the query $\boldsymbol{Q}_c \in \mathbb{R}^{1\times d}$, and flatten $\boldsymbol{F}_v$ into $N$ token features to form the key $\boldsymbol{K}_v \in \mathbb{R}^{N\times d}$ and value $\boldsymbol{V}_v \in \mathbb{R}^{N\times d}$. The clinical-guided context $\boldsymbol{z}$ is computed as:

$$\boldsymbol{z} = \mathcal{A}(\boldsymbol{f}_c, \boldsymbol{F}_v) = \mathrm{Softmax}\big(\boldsymbol{Q}_c {\boldsymbol{K}_v}^{\top}/\sqrt{d}\big)V_v, \tag{1}$$

where the Softmax operation is applied over the $N$ tokens. To avoid over-reliance on clinical variables, we further introduce an adaptive gating unit to balance the original and clinical-modulated features, as follows:

$$\hat{\boldsymbol{z}} = \mathrm{LN}(G \odot \boldsymbol{z} + (1-G) \odot \boldsymbol{f}_v), \quad G = \sigma(\boldsymbol{W}_G[\boldsymbol{f}_v \| \boldsymbol{z}]), \tag{2}$$

where $G \in \mathbb{R}^d$ is a channel-wise gating vector and $W_G \in \mathbb{R}^{d\times 2d}$ is a learnable matrix. $\odot$ denotes element-wise multiplication, $[\cdot \| \cdot]$ represents concatenation, and $\sigma(\cdot)$ is the sigmoid function. The resulting $\hat{\boldsymbol{z}}$ not only encodes the anatomical and functional information in mpMRI but also incorporates commonly used clinical risk factors such as PSA and age, thereby serving as a robust multimodal basis for subsequent knowledge learning and GGG prediction.

### 2.2 Knowledge-Guided Prototype Alignment (KGPA)

Although the recalibrated representation $\hat{\boldsymbol{z}}$ incorporates multi-modal cues, optimizing it only with categorical supervision still treats the GGG grades as flat and independent classes, overlooking their intrinsic severity order (GGG $\in \{1,2,3,4,5\}$). As a result, the learned embeddings may become scattered and overlap across adjacent grades. To address this, KGPA leverages radiology reports to construct grade-specific semantic anchors $\mathcal{P} = \{\boldsymbol{p}_k\}_{k=1}^{K}$ with $K = 5$. These anchors act as report-driven prototypes to regularize the latent feature space and promote ordinal-consistent representations.

**LLM-based Expert Knowledge Extraction.** To incorporate expert knowledge from radiology reports and clinical guidelines, we use an LLM (i.e., GPT-5.1) as an offline knowledge extractor. Specifically, we prompt the LLM to aggregate and summarize recurrent radiological findings from a large corpus of mpMRI reports for each grade group, producing a standardized population-level phenotype description $T_k$ for the $k$-th grade group. To improve clinical validity and consistency, established clinical guidelines are incorporated into the prompt as structural constraints (e.g., a unified template specifying required fields and terminology). The resulting descriptions provide a clinically grounded semantic basis for constructing ordinal prototypes.

**Ordinal-Aware Learnable Prototypes**. To bridge report-derived general knowledge with patient-specific mpMRI, we design a learnable prototype bank $\boldsymbol{E} = \{\boldsymbol{e}_k \in \mathbb{R}^{d_{emb}}\}_{k=1}^{K}$, where each prototype $\boldsymbol{e}_k$ is initialized from the embedding of $T_k$ and is subsequently refined during training. To encode the natural progression of disease severity in GGG (from 1 to 5), we assign each grade an ordinal positional embedding $\boldsymbol{o}_k \in \mathbb{R}^{\boldsymbol{d_{ord}}}$, within the range [-1, 1]. The ordinal-aware representation $\tilde{\boldsymbol{e}}_k$ is computed as:

$$\tilde{\boldsymbol{e}}_k = \boldsymbol{W}_F[\boldsymbol{e}_k \| \boldsymbol{o}_k] \in \mathbb{R}^d, \quad \boldsymbol{p}_k = \Phi_{pro}(\tilde{\boldsymbol{e}}_k) / \left\| \Phi_{pro}(\tilde{\boldsymbol{e}}_k) \right\|_2 \in \mathbb{R}^d, \tag{3}$$

where $\tilde{\boldsymbol{e}}_k$ is the fused prototype representation, $\boldsymbol{W}_F$ is a learnable linear projection layer, and $\Phi_{pro}(\cdot)$ is a mapping network. As training progresses, these anchors are refined via the following prototype contrastive alignment and gradually evolve into grade-specific centers in the embedding space.

**Prototype Contrastive Alignment.** To align patient-specific features $\hat{\boldsymbol{z}}$ from CCM with the knowledge anchors, we adopt the prototype contrastive loss $\mathcal{L}_{align}$ as follows:

$$\mathcal{L}_{align} = -\log \frac{\exp\left(\langle \boldsymbol{v}_{proj}, \boldsymbol{p}_y \rangle / \tau\right)}{\sum_{k=1}^{K} \exp\left(\langle \boldsymbol{v}_{proj}, \boldsymbol{p}_k \rangle / \tau\right)}, \tag{4}$$

where $\boldsymbol{v}_{proj}$ represents the projected features of $\hat{\boldsymbol{z}}$, $y \in \{1, \dots, K\}$ is the ground-truth label (thus $\boldsymbol{p}_y$ denotes the corresponding positive anchor). This constraint encourages an ordinal topology in the feature space, thereby faithfully mirroring the progression of cancer severity, ranging from mild (GGG=1) to advanced pathological stages (GGG=5).

### 2.3 Hierarchical Ordinal-aware Constraints (HOC)

**Pattern Decoupling and Prediction.** The GGG is defined by the combination of primary and secondary Gleason patterns according to ISUP standards. Inspired by this, the HOC module transforms conventional flat classification of GGG into a hierarchical learning process. To achieve this, we first decouple GGG prediction into primary and secondary pattern prediction using a dual-stream network. Given $\hat{\mathbf{z}}$ from CCM, the dual-stream network employs two independent pattern prediction layers, denoted as $H_p(\cdot)$ and $H_s(\cdot)$, to predict the probability distributions of the primary pattern $\hat{y}_p$ and the secondary pattern $\hat{y}_s$, respectively, as follows:

$$\hat{y}_p = \text{Softmax}\left(H_p(\hat{\mathbf{z}})\right), \quad \hat{y}_s = \text{Softmax}\left(H_s(\hat{\mathbf{z}})\right), \tag{5}$$

This process encourages the model to learn discriminative and fine-grained visual cues related to the dominant and subordinate pathological patterns, rather than simply memorizing global labels.

**Differentiable Bio-logic Mapping (DBML).** DBML maps the decoupled predictions to the final GGG distribution. We first compute the joint probability matrix $\boldsymbol{J} \in \mathbb{R}^{3\times3}$ as $\boldsymbol{J} = \hat{y}_p \otimes \hat{y}_s$, where $\otimes$ signifies the outer product. Each element $\boldsymbol{J}_{ij}$ corresponds to the joint probability of the primary pattern $i$ and the secondary pattern $j$ (where $i, j \in \{3,4,5\}$). Subsequently, we calculate the GGG distribution as $\hat{y}_{ggg} = \text{vec}(\boldsymbol{J}) \cdot \boldsymbol{W}_m$, where $\boldsymbol{W}_m$ is a frozen sparse matrix encoding the ISUP mapping logic. This differentiable operation enables end-to-end structured learning by propagating gradients from the final GGG supervision back to the decoupled pattern predictors in the dual-stream network, thus accurately reflecting the clinical grading process.

**Hierarchical Ordinal Loss.** To supervise the ordinal and hierarchical prediction of GGG, we develop the hierarchical ordinal loss, which consists of a pattern-level loss $\mathcal{L}_{pat}$, a grade-level loss $\mathcal{L}_{ggg}$, and a composite ordinal loss $\mathcal{L}_{ord}$. Specifically, the pattern-level loss aims to provide fine-grained supervision for the decoupled primary and secondary Gleason pattern predictions, formulated as:

$$\mathcal{L}_{pat} = CE\left(\hat{y}_p, y_p\right) + CE(\hat{y}_s, y_s), \tag{6}$$

where $CE(\cdot)$ represents the cross-entropy loss.

The grade-level loss applies an ordinal-sensitive regression loss [22] to supervise the final prediction of GGG score as:

$$\mathcal{L}_{ggg} = -\frac{1}{K-1}\sum_{j=1}^{K-1}\left[\mathbb{I}(y > j)\log\sigma\left(f_j\right) + (1 - \mathbb{I}(y > j))\log(1 - \sigma(f_j))\right], \tag{7}$$

where $K$ denotes the number of Grade Groups, $y \in \{1, \dots, K\}$ is the ground-truth GGG label, $j$ indexes the $K-1$ ordinal thresholds, and $\sigma(\cdot)$ is the sigmoid function.

The composite ordinal loss is adopted to adaptively penalize rank violations as:

$$\mathcal{L}_{ord} = \frac{1}{(K-1)^2}\sum_{k=1}^{K}\hat{y}_{ggg}[k]\cdot(k-y)^2, \tag{8}$$

where $\hat{y}_{ggg}[k]$ denotes the predicted probability on grade $k$. Notably, the quadratic term $(k-y)^2$ makes the penalty distance-aware, assigning larger costs to farther rank errors (e.g., $|k-y| = 4$ incurs16 times the penalty of $|k-y| = 1$).

Finally, the total objective is the weighted sum of the above losses, formulated as:

$$\mathcal{L}_{total} = \mathcal{L}_{pat} + \lambda_{ggg}\mathcal{L}_{ggg} + \lambda_{ord}\mathcal{L}_{ord} + \lambda_{align}\mathcal{L}_{align}. \quad (9)$$

where $\lambda_{ggg}$, $\lambda_{ord}$, and $\lambda_{align}$ are the hyperparameters to balance these terms.

# 3 Experiments and Results

## 3.1 Experimental Settings

**Datasets.** We trained and evaluated KOAL on a public PI-CAI dataset [23] of 643 patients and an in-house dataset of 1002 patients. For each patient, we collected mpMRI sequences, crucial clinical variables, and ground truth GGG scores with primary and secondary Gleason patterns. All MRI sequences were resampled to 20×320×320 and cropped into 20×96×96 tumor-centered volumes to reduce background redundancy.
**Evaluation Metrics.** We employ the Quadratic Weighted Kappa (QWK) as the primary evaluation metric. We also report the overall Accuracy (ACC) and the Area Under the Curve (AUC) metrics to evaluate discriminative capability across different grades.
**Implementation Details.** All experiments were implemented in PyTorch on a RTX 4090 GPU (24GB). The model was trained for 200 epochs (batch size 16) using the AdamW optimizer with an initial learning rate of 1e-4. The hyper-parameters $\lambda_{GGG}$, $\lambda_{ord}$, and $\lambda_{align}$ in Eq. (9) are empirically set to 1.0, 0.5, and 0.1, respectively.

## 3.2 Comparison with State-of-the-Art Methods

To validate our effectiveness, we compare KOAL with (1) image-only methods, including PON [16], CORAL [22], AOR-DR [24], and SupCon [25]; and (2) multi-modal methods, including HyperFusion [9], PPAL [19], DSPA-MIL [26], XCoOp [18], and DAFT [17]. The results are presented in Table 1 and Fig. 2.

**Table 1.** Quantitative comparison with nine state-of-the-art methods on the public PI-CAI dataset and the private In-House dataset. I: Imaging modality (mpMRI); T: Tabular clinical variables (e.g., PSA, age). The best results are highlighted in **bold**, and the second-best are underlined.

| Methods | M | | PI-CAI | | | In-House | | |
|---|---|---|---|---|---|---|---|---|
| | I | T | QWK(%) | AUC(%) | ACC(%) | QWK(%) | AUC(%) | ACC(%) |
| PON [16] | √ | | 65.42 | 76.62 | 47.42 | 62.60 | 68.31 | 51.66 |
| CORAL [22] | √ | | 77.06 | 84.18 | 56.70 | 60.11 | 73.23 | 42.38 |
| AOR-DR [24] | √ | | 74.05 | 81.81 | 57.14 | 65.80 | 80.57 | 56.70 |
| SupCon [25] | √ | | 79.70 | 89.42 | 72.16 | 67.20 | 83.92 | 64.12 |
| HyperFusion [9] | √ | √ | 82.66 | 83.82 | 60.82 | 77.96 | 62.96 | 59.60 |
| PPAL [19] | √ | √ | 82.64 | **92.07** | 69.41 | 64.62 | 82.44 | 57.25 |
| DSPA-MIL [26] | √ | √ | 76.19 | 85.61 | 70.10 | 53.49 | 76.70 | 52.98 |
| XCoOp [18] | √ | √ | 78.78 | 88.90 | 71.13 | 75.93 | 86.18 | 67.91 |
| DAFT [17] | √ | √ | 83.28 | 91.32 | 75.26 | 79.17 | **89.42** | 70.99 |
| KOAL (Ours) | √ | √ | **88.99** | 91.93 | **81.63** | **88.19** | 89.38 | **75.38** |

**Quantitative Results.** On PI-CAI, KOAL achieves the highest QWK (88.99%) and ACC (81.63%), alongside a 91.93% AUC. DAFT ranks second in QWK/ACC, while PPAL obtains a marginally higher AUC. To evaluate model robustness against clinical heterogeneity, we further validate on an In-House dataset with greater inter-patient var-iability. While baseline performances decline substantially on this dataset (e.g., PPAL's QWK decreases from 82.64% to 64.62%), KOAL maintains robust performance, yielding 88.19% QWK, 75.38% ACC, and 89.38% AUC. These results underscore KOAL's superior generalizability for clinical GGG prediction across varying data distributions.

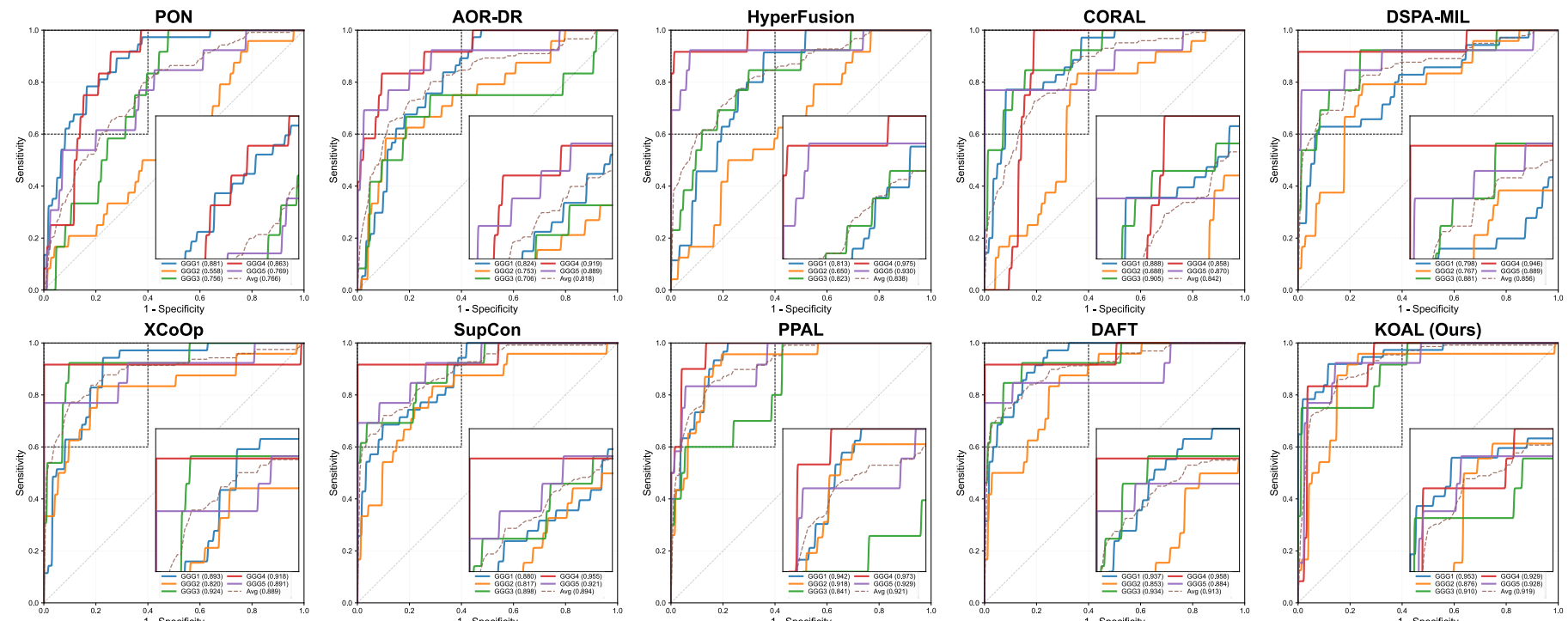

**Fig. 2.** Comparison of ROC curves between KOAL and competing methods on PI-CAI dataset.

### 3.3 Ablation Study

We conduct ablation studies on PI-CAI and the In-House datasets. As shown in Table 2, the mpMRI-only baseline performs the worst. CCM module yields substantial gains (↑10.27% QWK on PI-CAI; ↑12.82% on In-House). KGPA module further boosts performance and derives report-derived semantic anchors to align patient-specific mpMRI representations. While standard ordinal loss (without primary/secondary decoupling) highlights the benefit of ordinal supervision, integrating the HOC module achieves optimal results, confirming the advantage of explicitly modeling the intrinsic compositional order of primary and secondary Gleason patterns in GGG prediction.

**Table 2.** Ablation study on the PI-CAI dataset and the In-House dataset. M: MpMRI (Baseline); C: CCM module; K: KGPA module; H: HOC module; O: standard ordinal loss treating GGG as a single ordered label (without primary/secondary decoupling).

| Variants | | | | | PI-CAI | | | In-House | | |
|---|---|---|---|---|---|---|---|---|---|---|
| M | C | K | O | H | QWK(%) | AUC(%) | ACC(%) | QWK(%) | AUC(%) | ACC(%) |
| √ | | | | | 65.51 | 77.09 | 39.80 | 56.07 | 72.51 | 42.38 |
| √ | √ | | | | 75.78 | 78.78 | 44.90 | 68.89 | 82.06 | 58.28 |
| √ | √ | √ | | | 83.67 | 88.29 | 68.37 | 74.17 | 83.46 | 57.62 |
| √ | √ | √ | √ | | 85.04 | 89.83 | 74.23 | 84.60 | 84.74 | 68.90 |
| √ | √ | √ | | √ | **88.99** | **91.93** | **81.63** | **88.19** | **89.38** | **75.38** |

## 4 Conclusion

In this paper, we propose KOAL, a novel knowledge-driven and ordinal-aware framework for non-invasive GGG prediction using mpMRI, clinical variables, and radiology reports. Specifically, we introduce the CCM module, which modulates mpMRI repre-sentations based on patient-specific clinical variables. Meanwhile, the KGPA module leverages semantic anchors extracted from radiology reports by a LLM to align with the image representations. Finally, the HOC module models the hierarchical structure of GGG via pattern decoupling and DBML mapping. Experimental results on two prostate cancer datasets validate that our method achieves state-of-the-art performance, suggesting that KOAL is a promising tool for non-invasive prostate cancer assessment.